\UseRawInputEncoding

\documentclass[journal]{IEEEtran}
\usepackage{epsfig} 
\usepackage{epstopdf}
\usepackage{amsfonts}
\usepackage{amsmath}
\usepackage{amssymb}
\usepackage{graphicx}
\usepackage{algpseudocode}
\usepackage{array}
\usepackage{booktabs}
\usepackage{makecell}
\usepackage{cite}
\usepackage{color}
\usepackage{soul}
\usepackage{subfigure}

\soulregister\ref7
\soulregister\cite7

\usepackage[linesnumbered,ruled,vlined]{algorithm2e}
\usepackage{algpseudocode}
\SetKwBlock{Iteration}{Iteration }{end}

\newtheorem{definition}{Definition}

\begin{document}

\title{Coverage Optimization of Camera Network for Continuous Deformable Object}

\author{Chang~Li,
        Xi~Chen,~\IEEEmembership{Member,~IEEE,}
        Li~Chai,~\IEEEmembership{Member,~IEEE}
\thanks{*This work was supported under the National Natural Science Foundation
of China under Grant 62173259.}

\thanks{Chang Li is with the School of Information Science and Engineering, Wuhan University of Science and Technology, Hubei, 430081 China (e-mail: lichang94@wust.edu.cn).}
\thanks{Xi Chen is with the Engineering Research Center of Metallurgical Automation and Measurement Technology, Wuhan University of Science and Technology, Hubei 430081, China (e-mail: chenxi\_99@wust.edu.cn) (corresponding author).}
\thanks{Li Chai is with the College of Control Science and Engineering, Zhejiang University, Hangzhou, 310027, China (chaili@zju.edu.cn).}
}



\maketitle

\begin{abstract}
In this paper, a deformable object is considered for cameras deployment with the aim of visual coverage. The object contour is discretized into sampled points as meshes, and the deformation is represented as continuous trajectories for the sampled points. To reduce the computational complexity, some feature points are carefully selected representing the continuous deformation process, and the visual coverage for the deformable object is transferred to cover the specific feature points. In particular, the vertexes of a rectangle that can contain the entire deformation trajectory of every sampled point on the object contour are chosen as the feature points. An improved wolf pack algorithm is then proposed to solve the optimization problem. Finally, simulation results are given to demonstrate the effectiveness of the proposed deployment method of camera network.
\end{abstract}


\section{Introduction}

\IEEEPARstart{C}{amera} network is widely used to fulfill various visual tasks in industry, such as surveillance \cite{Ahmed2021,Du2019surveillance}, industrial inspection \cite{Mavrinac2015inspection}, motion capture \cite{Park2015MotionCapture}, 3-D reconstruction \cite{Moemen20213D,Zhao20193d} and so on. Among these implementations, visual coverage is a crucial problem, whose objective is to achieve environmental perception via deploying the visual sensors in 2-D \cite{Zhao2014,Abbasi2017,Parapari2016} or 3-D areas \cite{zhang20153,leitmech,zhang2018visual}. The main approach for visual coverage follows three steps. The first step is to partition the object to be covered into many meshes, and a sampling point is selected to represent each partitioned mesh. The second step is to analyze the visual performance of the camera network on each sampled point, with considering the visual criterion of camera, i.e., resolution, field of view and focus etc.. The third step is to build the cost function describing the total visual coverage characteristics of the camera network for the object, and a suitable optimization algorithm is designed to solve the optimization problem. It is noted that the computational complexity of the optimization is closely related to the size of the sampled points for the object in step one. The bigger the size of the sampled points is, the longer computational cost of the optimization needs.

Most of visual coverage studies focus on the static object, meaning that the size and the postures of the partitioned sampled points or meshes of the object is fixed since initial time. In some issues, the contour of the object to be covered may change over time. For instance, for the parts in machining process, it needs to deploy the cameras for monitoring the whole machining process. This kind of problem can be regarded as visual coverage for deformable object, which has been discussed in \cite{Herguedas2019,Schacter2018,Neger2018} via dynamical deploying the cameras according to the deformation process. Due to the field of view for the camera is always wide, the static deployment of cameras is sufficient for visual coverage on deformable object. A straightforward method of static deployment of cameras for deformable object is to follow the corresponding approach for static object in \cite{zhang20153,leitmech,zhang2018visual}. Specifically, with partitioning the deformation trajectories of the sampled points on the object into many discrete points, the objective of the visual sensors deployment is to cover all the sampled points on the object along the entire deformation process. However, a serious problem of this method is that the optimization will take a incredibly long time with the huge size of the total sampled points.

In order to reduce the computational cost of the optimization, a novel approach is proposed in this paper for visual coverage on deformable object. Specifically, a rectangle area is built for each deformation trajectory of every sampled point on the object contour. The entire deformation trajectory of of each point is completely contained by its rectangle. Then, the vertexes of the rectangles are selected as the feature points representing the continuous  deformation trajectories. The visual coverage of deformable object is achieved if the selected feature points are covered by the networked cameras. By this approach, the optimization speed can be significantly accelerated. In addition, an improved wolf pack algorithm with faster convergence rate and better performance than the standard wolf pack algorithm, is designed to solve the formulated optimization problem.

The remaining parts of this paper are organized as follows. Section II introduces the models of the deformable object and the cameras, as well as the problem statement. In Section III, the procedure of the feature points selection for deformation process is stated, and the cost function is established accordingly. An improved wolf pack algorithm is proposed in Section IV to solve the deployment optimization. In Section V, some numerical examples are given to illustrate the effectiveness of the proposed method. Finally, the paper is concluded in Section VI.

\emph{Notations:} Let $\mathbb{R}^n$ denote $n$-dimensional Euclidean space, $\mathbb{N}^+$ represents a positive integer, and $SO(2)$ denotes Special Orthogonal Group in 2-D planar.

\section{Modelling and Problem Statement}\label{sec:II}

In this section, the models of deformable object and camera network are introduced, respectively, and the problem statement is then presented.

\subsection{Deformable Object Model}

A 2-D object model to be covered is introduced in world framework $\mathcal{F}_w$ at first. Without loss of generality, the object contour is composed of sampled segments of length being smaller than a threshold $L_s$. Let $\tau_k$ denote the $k^{th}$ segment ($k=1,\cdots,K$) with $K\in\mathbb{N}^+$ being the total number of the segments on the object contour. Denote $\mathbf{p}_k=[\mathbf{s}_k^\mathrm{T} ~\rho_k]^\mathrm{T}\in\mathbb{R}^2\times[0,2\pi)$ to be the coordinate of the centroid of $\tau_k$, where $\mathbf{s}_k=[x_k~y_k]^\mathrm{T}\in\mathbb{R}^2$ is the position component and $\rho_k\in[0,2\pi)$ is the orientation component of the front face normal direction of $\tau_k$ in $\mathcal{F}_w$. Then, the normal vector of $\tau_k$ can be expressed as $\mathbf{n}_k=[\cos\rho_k,\sin\rho_k]$.

The 2-D object contour will be deformed under the external force $u$. Before modelling the deformation process of the object contour, two assumptions are made as follows.

	\newtheorem{assumption}{Assumption}
	\begin{assumption}\label{ass:finite time force}
		The known external force $u$ is added to the object on a finite time interval $[t_0,t_s]$ with $t_s>t_0>0$. The object model before time instant $t_0$ is known for visual coverage in advance, i.e., $\mathbf{p}_1(t),\cdots,\mathbf{p}_K(t)$ for $t<t_0$.
	\end{assumption}
	
	\begin{assumption}\label{ass:small deformation}
		The value of external force $u$ is small such that during the time interval $[t_0,t_s]$ the deformation of the object is small. Moreover, the continuity of the object contour is preserved, in other words, the object will not be split into separated parts.
	\end{assumption}

Under Assumptions \ref{ass:finite time force} and \ref{ass:small deformation}, some facts can be induced for deformable object. First, the model of the object is fixed before $t_0$ and after $t_s$. The coordinates of the sampled points on the contour can be obtained by some instruments, i.e., $\mathbf{p}_1(t),\cdots,\mathbf{p}_K(t)$ for $t<t_0$ and $t>t_s$. Second, the position information for each sampled point on the object contour can be calculated at any time $t\in[t_0,t_s]$ if the potential associated to the external force is known \cite{Cohen2010}. Third, due to the deformation is small, the change of orientation for each sampled point can be assumed to be continuous and uniform on time interval $[t_0,t_s]$. To this end, the deformable object contour can be modelled as a time-varying manifold $\Omega(t)=\{\mathbf{p}_1(t),\cdots,\mathbf{p}_K(t)\}$ and are always available for visual coverage.

\subsection{Camera Model}

\begin{figure}[!t]
		\vspace{3pt}
		\centering
		\includegraphics[scale=0.7]{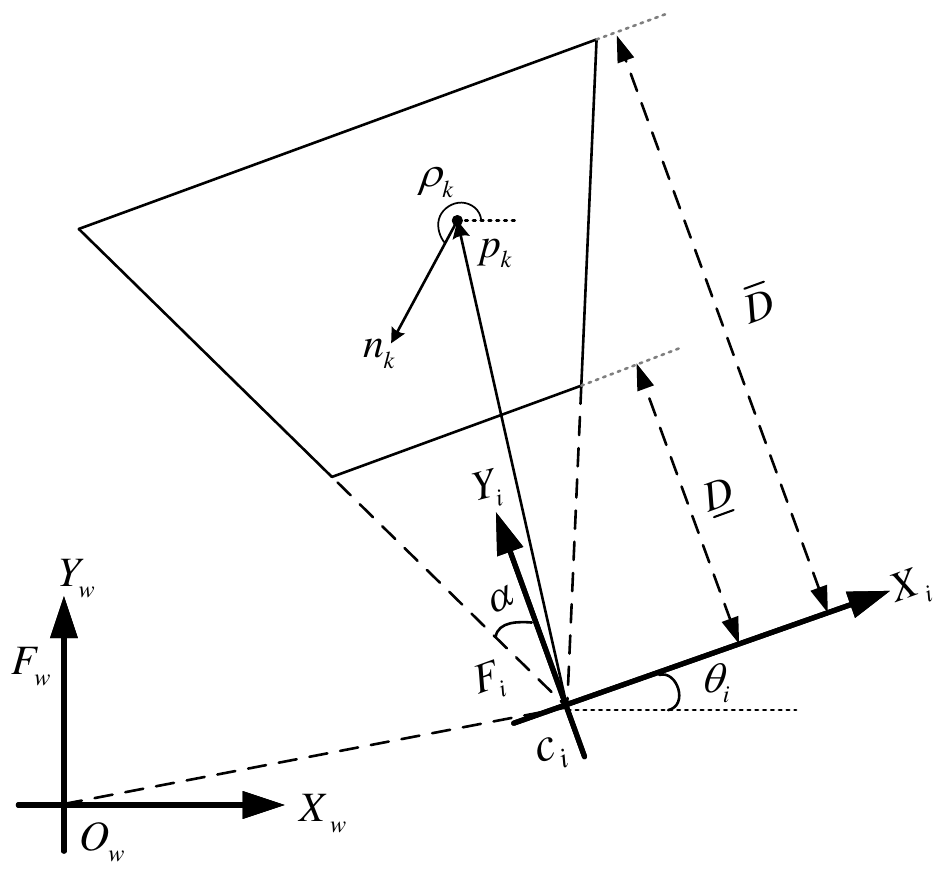}
		\caption{Camera model.}
		\label{fig:camrea_model}
		\vspace{-1pt}
	\end{figure}

A group of $N\in\mathbb{N}^+$ cameras is employed to perform visual coverage task for deformable object. For the $i^{th}$ camera ($i=1,\cdots,N$), its geometric model is formed as an isosceles trapezoid which represents the field of view (FOV), depicted in Fig. \ref{fig:camrea_model}. The $i^{th}$ camera is located at the intersection of the legs of the isosceles trapezoid with position $\mathbf{v}_i=[x^c_i~y^c_i]\in\mathbb{R}^2$ and the orientation of the central line $\theta_i\in [0,2\pi)$ in world frame $\mathcal{F}_w$. The position $\mathbf{v}_i$ together with the orientation $\theta_i$ form the configuration of the $i^{th}$ camera as $\mathbf{c}_i=[\mathbf{v}_i^\mathrm{T}~\theta_i]^\mathrm{T}$. Let $\Lambda=\{\mathbf{c}_1,\cdots,\mathbf{c}_N\}$ be the set of configurations for the camera network. The distances from the bases of the isosceles trapezoid to the camera are $\underline{D}$ and $\overline{D}$, respectively, and the field angle between the legs of the isosceles trapezoid is $2\alpha$. Moreover, a local frame $\mathcal{F}_i$ is defined on the $i^{th}$ camera whose origin coincides with the camera. The $\mathbf{Y}_i$ axis is aligned with the central line of the isosceles trapezoid pointing to the FOV, and the $\mathbf{X}_i$ axis is then constructed such that the frame $\mathcal{F}_i$ formed by right-hand rule.

For each sampled point on the object contour, its position $\mathbf{s}_k$ in $\mathcal{F}_w$ can be transformed into the coordination $\mathbf{s}_k^{c_i}=[x_k^{c_i}~y_k^{c_i}]\in\mathbb{R}^2$ in $\mathcal{F}_i$ by
\begin{equation}\label{eq:coordination transformation}
\mathbf{s}_k^{c_i}=R_i(\theta_i)(\mathbf{s}_k-\mathbf{v}_i)
\end{equation}
with $R_i(\theta_i)\in SO(2)$ being the rotation matrix as
\begin{equation}
R_i(\theta_i)=\left[\begin{array}{cc}
\cos\theta_i & \sin\theta_i\\
-\sin\theta_i & \cos\theta_i
\end{array}\right].
\end{equation}

\subsection{Problem Statement}

For a directional sampled point on object contour $\mathbf{p}_k$, it can be visible by a camera $\mathbf{c}_i$ if for one hand $\mathbf{p}_k$ falls in the FOV of $\mathbf{c}_i$, and for the other hand, the relative angle between the front face normal direction of $\tau_k$ and the vector from $\mathbf{c}_i$ to $\mathbf{p}_k$ is in $(\pi/2,\pi]$. The definition of visible sampled point is given as follows,
\begin{definition}
	For sampled point on deformable object contour $\mathbf{p}_k(t), k=1,\cdots,K$. It is said to be visible by the camera $\mathbf{c}_i$ if the following conditions are satisfied
\begin{align}
\underline{D}\leq y_k^{c_i}\leq \overline{D} \\
\arccos\frac{\overrightarrow{\mathbf{c}_i\mathbf{p}_k}[-\sin\theta_i,\cos\theta_i]^\mathrm{T}}{||\overrightarrow{\mathbf{c}_i\mathbf{p}_k}||\cdot||[-\sin\theta_i,\cos\theta_i]||}\leq \alpha \\
\arccos\frac{\overrightarrow{\mathbf{c}_i\mathbf{p}_k}[\cos\rho_k(t),\sin\rho_k(t)]^\mathrm{T}}{||\overrightarrow{\mathbf{c}_i\mathbf{p}_k}||\cdot||[\cos\rho_k(t),\sin\rho_k(t)]||}> \frac{\pi}{2}
\centering
\end{align}
with $\overrightarrow{\mathbf{c}_i\mathbf{p}_k}=[x_k-x_i^c(t),y_k-y_i^c(t)]$.
	\end{definition}

Given a camera network $\Lambda$, the task of visual coverage for deformable object is to find a set of configurations of camera network, such that for the deformable object $\Omega(t)$, the sampled points on the object contour are visible by the camera network over the time.

\section{Visual Coverage For Deformations}

This section aims at selecting some 'feature points' representing the continuous deformation process, and the cost function is then established to characterise the coverage performance of the total selected feature points.

\subsection{Feature Points Selection}

Denote a function $P_k(t), k=1,\cdots,K$ to represent the trajectory of the deformation process for point $\mathbf{p}_k$ on the object contour. For each trajectory $P_k(t)$ over time interval $[t_0,t_s]$, descretize it into $M$ sampled points as $\mathbf{p}_{k1},\cdots,\mathbf{p}_{kM}$ uniformly with $\mathbf{p}_{k1}=\mathbf{p}_k(t_0)$ and $\mathbf{p}_{kM}=\mathbf{p}_k(t_s)$. More specifically, the distances between any adjacent points on each trajectory, i.e., $\mathbf{p}_{kj},\mathbf{p}_{k(j+1)},j=1,\cdots,M-1$, are same and smaller than a threshold $d$. For simplicity of presentation, take a specific trajectory $P_{k'}(t)$ with $k'\in\{1,\cdots,K\}$ as an example. The positions of the sampled points on $P_{k'}(t)$ are represented by $\mathbf{s}_{k'1},\cdots,\mathbf{s}_{k'M}$ with $\mathbf{s}_{k'm}=[x_{k'm}~y_{k'm}]^\mathrm{T}\in\mathbb{R}^2$ for $m=1,\cdots,M$. The orientation of the front face normal direction for each sampled point $\mathbf{p}_{k'm}$ is defined as
\begin{equation}\label{eq:orientation of sampled point}
\rho_{k'm}=\rho_{k'1}+\frac{\rho_{k'M}-\rho_{k'1}}{M-1}(m-1), ~m=1,\cdots,M.
\end{equation}
Then, the coordination of each sampled point is obtained as $\mathbf{p}_{k'm}=[\mathbf{s}_{k'm}^\mathrm{T}~\rho_{k'm}]^\mathrm{T}\in\mathbb{R}^2\times[0,2\pi)$ for $m=1,\cdots,M$.

A convex hull is constructed that contains the trajectory of the whole deformation process for every point on the object contour. Without loss of generality, a rectangle denoted as $Rec_{k'}$ is established in this paper as a special case for the convex hull. More specifically, the vertexes of $Rec_{k'}$ are denoted respectively by $\nu_{k'1},\nu_{k'2},\nu_{k'3}, \nu_{k'4}$, whose positions are $[x^{k'}_{\min},y^{k'}_{\min}]$, $[x^{k'}_{\min},y^{k'}_{\max}]$, $[x^{k'}_{\max},y^{k'}_{\min}]$ and $[x^{k'}_{\max},y^{k'}_{\max}]$ with
\begin{align}
\label{eq:position 1 of vertex}x^{k'}_{\min}&=\min\{x_{k'1},\cdots,x_{k'M}\} \\
\label{eq:position 2 of vertex}x^{k'}_{\max}&=\max\{x_{k'1},\cdots,x_{k'M}\} \\
\label{eq:position 3 of vertex}y^{k'}_{\min}&=\min\{y_{k'1},\cdots,y_{k'M}\} \\
\label{eq:position 4 of vertex}y^{k'}_{\max}&=\max\{y_{k'1},\cdots,y_{k'M}\}.
\end{align}
In addition, the orientation of each vertex is defined as same as the orientation of the nearest sampled point on the trajectory to the vertex. Obviously, the trajectory of the sampled point $p_{k'}(t)$ is surrounded by the rectangle whose vertexes positions defined by (\ref{eq:position 1 of vertex})-(\ref{eq:position 4 of vertex}). Consequently, all the trajectories of the whole deformation process are contained by $K$ rectangles. The $4K$ vertexes of the rectangles are then selected as the 'feature points'. The coordinations of the feature points are denoted as $\mathbf{p}^\nu_j=[{\mathbf{s}_j^\nu}^\mathrm{T},\rho_j^\nu]\in\mathbb{R}^2\times[0,2\pi)$ for $j=1,\cdots,4K$, with $\mathbf{s}_j^\nu$ being the position component and $\rho_j^\nu$ being the orientation component.

The procedure for the feature points selection is demonstrated by Algorithm 1 .

\begin{algorithm}[!t]
		\SetAlgoLined
		\caption{Feature Points Selection Algorithm}
		
		\KwIn{$p_{k}(t_0),p_{k}(t_s),K,M$}
		\KwOut{$\mathbf{p}^\nu_j$}

		\Iteration{
			\For{$k=1;k \le K; k++$}{
			
		Denote function $P_{k}(t)$;\\
		Sampling $M$ points from $p_{k}(t_0)$ to $p_{k}(t_s)$;\\
		Obtain $\mathbf{p}_{km}=[\mathbf{s}_{km}^\mathrm{T}~\rho_{km}]^\mathrm{T},m=1,\cdots,M$;\\	
		Calculation operation by formula $(7)-(10)$;\\
		Obtain $\nu_{k1},\nu_{k2},\nu_{k3},\nu_{k4}$;\\
			
			\For{$m=1;m \le M; m++$}{
           Calculate $d_{k'm1}= \|\mathbf{s}_{km}\cdot\nu_{k1}\|$;\\
           Same operation for $d_{km2},d_{km3},d_{km4}$
           }
           Calculate $d_{min}^{km1}= \min\{d_{km1},\cdots,d_{km1}\}$;\\
           Same operation for $d_{min}^{km2},d_{min}^{km3},d_{min}^{km4}$;\\

           Find the index $m_{k1}$ corresponding to $d_{min}^{km1}$;\\
           Same operation for $m_{k2},m_{k3},m_{k4}$;\\

           Select $\rho_{k1}^\nu=\rho_{km_{k1}}$;\\
           Same operation for $\rho_{k2}^\nu,\rho_{k3}^\nu,\rho_{k4}^\nu$;\\
           Pick $\mathbf{s}_{4(k-1):4k}^\nu=[\nu_{k1},\nu_{k2},\nu_{k3},\nu_{k4}]$;\\
           Pick $\rho_{4(k-1):4k}^\nu=[\rho_{k1}^\nu,\rho_{k2}^\nu,\rho_{k3}^\nu,\rho_{k4}^\nu]$;\\
			}
			
			Obtain $\mathbf{p}^\nu_j \ for \ j=1,\cdots,4K$\;

		}
	\end{algorithm}

\subsection{Cost Function}

With the selected feature points in hand, the coverage optimization for the deformable object contour is transferred into the coverage optimization for the feature points. A binary function is given as follows to describe whether a feature point $\mathbf{p}_j^\nu, j=1,\cdots,4K$ is visible by a specific camera $\mathbf{c}_i, i=1,\cdots,N$
\begin{equation}
			C_s({\mathbf{c}_i},\mathbf{p}_{j}^\nu) =
			\begin{cases}
				{1} & \text{visible},\\
				{0} & \text{otherwise}.\\
			\end{cases}
			\label{eq:C_s}
		\end{equation}
Then, the cost function is constructed as
\begin{equation}\label{eq:cost function}
\mathcal{H}(\mathbf{c}_1,\cdots,\mathbf{c}_N,\mathbf{p}_1^\nu,\cdots,\mathbf{p}_{4K}^\nu)=\Sigma_{j=1}^N h_j(\mathbf{c}_1,\cdots,\mathbf{c}_N,\mathbf{p}_j^\nu)
\end{equation}
where
\begin{equation}
h_j(\mathbf{c}_1,\cdots,\mathbf{c}_N,\mathbf{p}_j^\nu)=\max\{C_s({\mathbf{c}_1},\mathbf{p}_{j}^\nu),\cdots,C_s({\mathbf{c}_N},\mathbf{p}_{j}^\nu)\}.
\end{equation}

Hence, the deployment optimization of networked cameras for visual coverage on deformable object is formulated as
\begin{align}\label{eq:optimization}
&\mathop {\arg\max }_{{\mathbf{c}_1},\cdots,{\mathbf{c}_N}}~\mathcal {H}(\mathbf{c}_1,\cdots,\mathbf{c}_N,\mathbf{p}_1^\nu,\cdots,\mathbf{p}_{4K}^\nu),\\
&s.t.~\mathbf{c}_i\in\Lambda.		
\end{align}

\section{Optimization Algorithm}

The wolf pack algorithm (WPA) as a metaheuristic algorithms is widely used to solve the discrete optimization problem due to its global convergence and computational robustness. However, the standard wolf pack algorithm possess no good  performance for multiple variables optimization, i.e., positions and orientations of cameras. Therefore, an improved wolf pack algorithm (IWPA) is proposed in this paper to solve the deployment optimization problem. An additional operation for optimization orientation is further added in the optimization process, such that IWPA has better convergence performance and faster convergence rate than WPA.

\begin{figure}[!t]
		\vspace{3pt}
		\centering
		\includegraphics[scale=0.86]{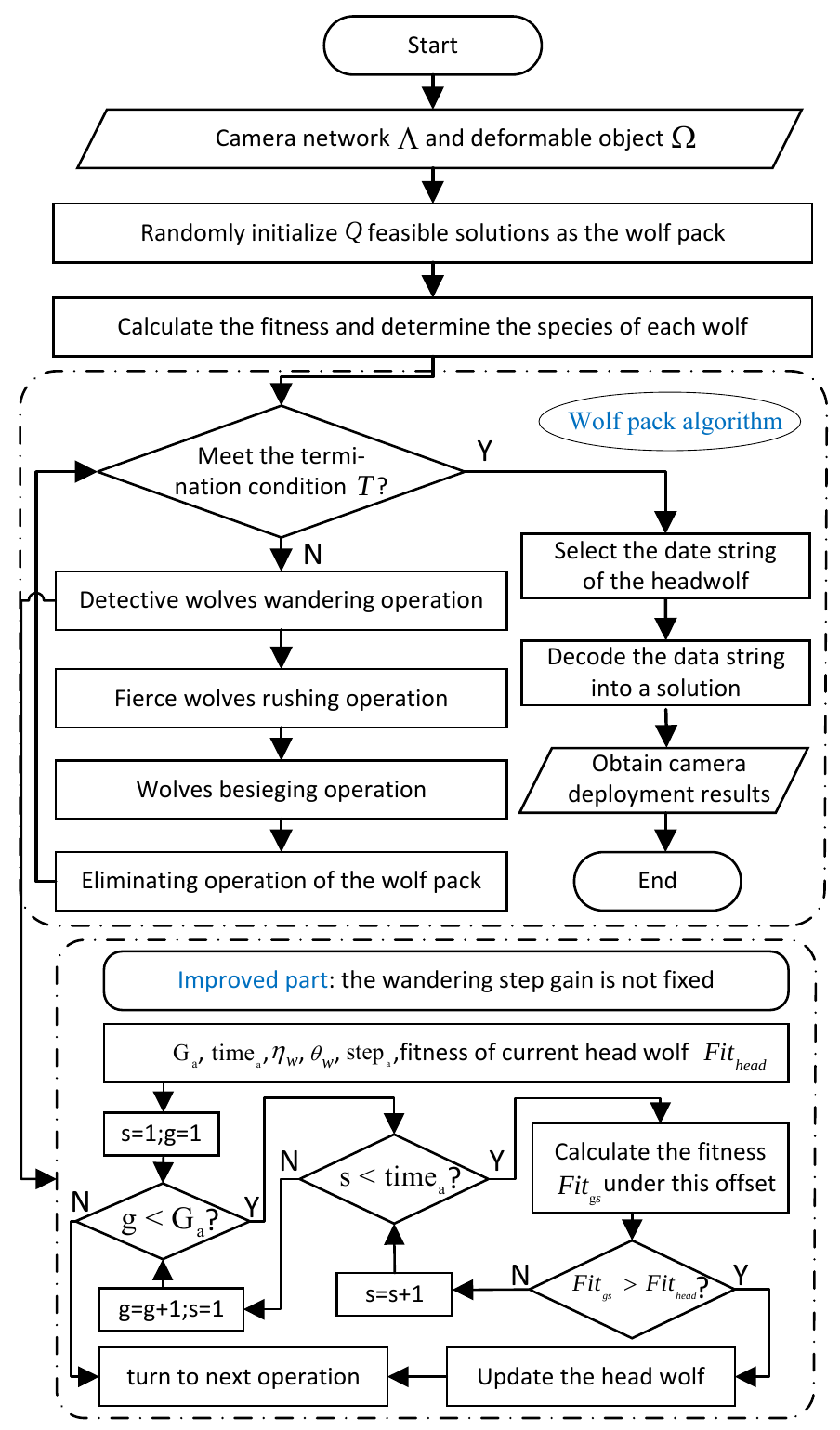}
		\caption{The flowchart of improved wolf pack algorithm.}
		\label{Fig:IWPA}
		\vspace{-1pt}
	\end{figure}

In the IWPA, each wolf represents a feasible solution which is encoded as a data string $\mathbf{WH}= \{c_1,\cdots,c_N\}$, and several wolves constitute the entire wolf pack as $\eth = \{\mathbf{WH}_1,\mathbf{WH}_2,\cdots,\mathbf{WH}_Q\}$ with $\mathbf{WH}_q$ ($q=1,2,\cdots,Q$) representing the $q^{th}$ wolf. The fitness of each wolf is described as the probability that the wolf searches for the prey, which can be quantified by the fitness function $Fit(\mathbf{WH})=\mathcal {H}(\mathbf{c}_1,\cdots,\mathbf{c}_N,\mathbf{p}_1^\nu,\cdots,\mathbf{p}_{4K}^\nu)$. The parameters of the IWPA are shown in Table. \ref{parameters_of_IWPA}. The procedure of the IWPA is described as follows and shown in Fig. \ref{Fig:IWPA}.

\begin{table}[!t]
		\renewcommand{\arraystretch}{1.3}
		\caption{Parameters of Improved Wolf Pack Algorithm}
		\label{parameters_of_IWPA}
		\centering
		\begin{tabular}{ll}
			\hline\hline
			\specialrule{0em}{0pt}{2pt}
			Parameter & Description \\
			\specialrule{0em}{2pt}{0pt}
			\hline
			\specialrule{0em}{2pt}{0pt}
			$Q\in\mathbb{N}^+$ & Size of wolf pack \\
			$L\in\mathbb{N}^+$ & Length of date string  \\
			$\Upsilon_{d}\in\mathbb{N}^+$  & Proportion of detective wolves \\
			$\Upsilon_{e}\in\mathbb{N}^+$  & Proportion of eliminated wolves \\
			$time_{a}\in\mathbb{N}^+$  & Maximum times of the wandering operation \\
			$\theta_{w} \in\mathbb{N}^+$ & Initial offset of wandering direction\\
			$G_{a}\in\mathbb{N}^+$ & The size of wandering step gain changes \\
			$step_{a} \in\mathbb{N}^+$ & Wandering step of detective wolves\\
			$\eta_{w} \in\mathbb{N}^+$ & Initial offset of Wandering step gain\\
			$time_{b}\in\mathbb{N}^+$ & Maximum times of the rushing operation\\
			$step_{b} \in\mathbb{N}^+$ & Rushing step of fierce wolves\\
			$D_{be}\in\mathbb{N}^+$ & Judgment distance of the besieging operation\\			
			$time_{c}\in\mathbb{N}^+$ & Maximum times of the besieging operation\\
			$\lambda_{c}\in\mathbb{N}^+$ & Besieging step gain of wolves\\
			$step_{c} \in\mathbb{N}^+$ & Besieging step of fierce wolves\\			
			$T \in\mathbb{N}^+$ & Algorithm termination condition \\
			\specialrule{0em}{2pt}{0pt}
			\hline\hline
		\end{tabular}
	\end{table}

\subsubsection{Initializing the wolf pack}
Calculate the fitness of each wolf in the wolf pack, the wolves are then classified in descending order of fitness. The wolf with the largest fitness is selected as the head wolf, the $\Upsilon_{d} \times Q$ wolves are selected as the detective wolves which fitness are slightly lower than the head wolf. The remaining wolves are all selected as fierce wolves. Suppose the prey is the optimal solution, each type of wolf does different operations to hunt for the prey.

\subsubsection{Detective wolves wandering operation}
The wandering operation is to allow the detective wolf to go hunting within the possible range to get a higher fitness. It can be understood as searching for a local optimal solution around the current suboptimal solutions. In each iteration, each detective wolf explores $time_{a}$ times around itself, the direction of each exploration is calculated by function $\sin(\cdot)$. In order to improve the effectiveness of the algorithm, the exploration range of detective wolf is from large to small. It is worth reminding that the wandering step $step_{a}$ needs to be set different values in position and orientation ($step_{ap}$ and $step_{ao}$).

Let $x^{t}_{ld}$ denote the current situation of the $l^{th}$ detective wolf in the $t^{th}$ iteration. $g$ and $s$ are natural numbers less than $G_{a}$ and $time_{a}$, respectively. The temporary wandering situation of this wolf in a certain direction and a certain wandering step can be expressed

 	\begin{equation}
 	\label{wandering operation}
		x^{t+1}_{ld}=x^{t}_{ld} + (1-\frac{g}{G_{a}}+\eta_{w}) \cdot step_{a} \cdot \sin(2\pi \cdot \frac{s}{time_{a}} + \theta_{w}).
	\end{equation}
Note that the terms $-\frac{g}{G_{a}}+\eta_{w}$ and $2\pi \cdot \frac{s}{time_{a}}$ are newly added in the IWPA comparing with WPA. Bringing in these two terms make the local optimum more likely to be found in this operation. With calculating the fitness of numerous temporary wandering situations, the optimized situation of the detective wolf is selected as the temporary situation with the greatest fitness. In the process of wandering operation, if the fitness of a detective wolf is higher than that of the head wolf, the head wolf would be updated and the original head wolf turns into a detective wolf.

\subsubsection{Fierce wolves rushing operation}
The rushing operation is to allow the fierce wolves to rush towards the head wolf at a larger pace. In order to distinguish the rushing operation and the besieging operation, a distance threshold $D_{be}$ is given. When the distance from the fierce wolf to the head wolf is greater than the distance $D_{be}$, the fierce wolf executes the rushing operation, otherwise the wolf executes the besieging operation. Each fierce wolf needs to rush to the distance $D_{be}$ away from the head wolf within the maximum times of rushing operation $time_{b}$.

Let $x^{t}_{lf}$ denote the solution of the $l^{th}$ fierce wolf in the $t^{th}$ iteration, and $x^{t}_{h}$ denotes the solution of current head wolf. The next solution of this fierce wolf can be expressed as

 	\begin{equation}
 	\label{rushing operation}
x^{t+1}_{lf}=x^{t}_{lf} + step_{b} \cdot \dfrac{(x^{t}_{h} - x^{t}_{lf})}{\mid x^{t}_{h} - x^{t}_{lf}\mid}.
	\end{equation}
		
In the process of rushing operation, if the fitness of a fierce wolf is higher than that of the head wolf, the head wolf would be updated.

\subsubsection{Wolves besieging operation}
Regarding the solution of the current head wolf as the prey, the besieging operation is to allow the fierce wolves and the detective wolves besiege the prey together at a small pace. Each wolf moves $time_{c}$ times towards the prey to search for the optimal solution.

Let $P^{t}$ denote the current prey in the $t^{th}$ iteration, and $x^{t}$ denote the solution of the $l^{th}$ wolf. The besieging operation of wolves can be expressed as

 	\begin{equation}
 	\label{besieging operation}
		x^{t+1}_{l}=x^{t}_{l} + \lambda_{c} \cdot step_{c} \cdot \mid P^{t} - x^{t}_{l} \mid.	
	\end{equation}
In the process of the besieging operation, if the fitness of any wolf except the head wolf is higher than that of the head wolf, the head wolf would be updated. 	

\subsubsection{Updating the wolf pack}
According to the principle of survival of the fittest, the wolf pack eliminates $\Upsilon_{e} \times Q$ wolves with low fitness and randomly generates an equal number of wolves to join the pack at the end of each iteration. The update mechanism makes the optimization algorithm less likely to fall into a local optimal solution.

\section{Numerical Simulation}
\begin{figure}[!t]
		\vspace{3pt}
		\centering
		\includegraphics[scale=0.40]{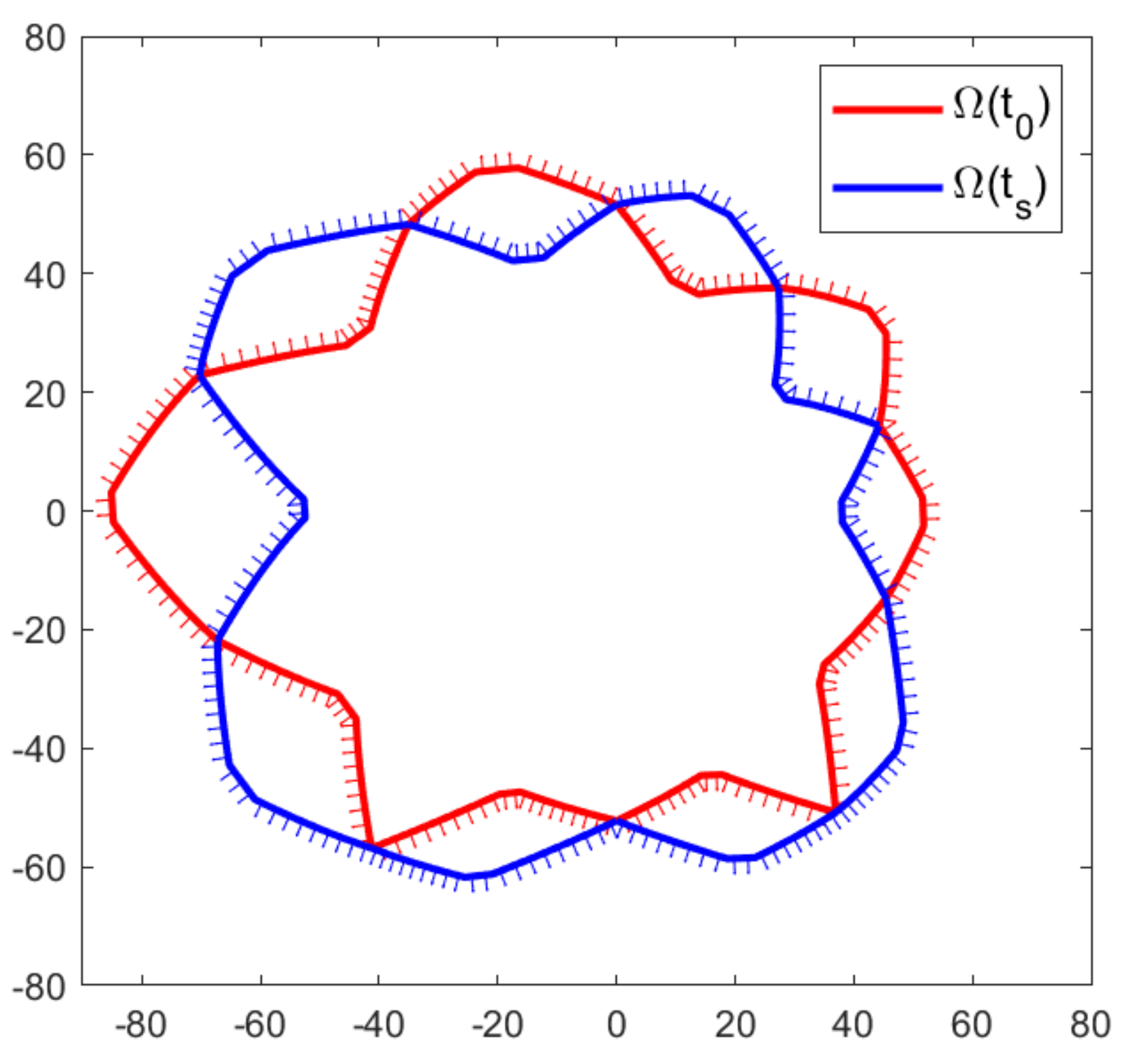}
		\caption{The deformable contour.}
		\label{Object contour}
		\vspace{-1pt}
	\end{figure}

In this section, numerical examples are given to verify the coverage performance of the proposed method. An object contour is randomly generated  consisting of $K=180$ sampled points. The object contour $\Omega$ is deformed by the external force from $t_0=0s$ with the object contour represented by red curves to $t_s=11s$ with the object contour represented by blue curves, as shown in Fig. \ref{Object contour}.

A group of $N=6$ cameras are given to be deployed to complete the coverage of the deformable object contour $\Omega(t)$. Half of the field angle of camera model is set as $\alpha=26^{\circ}$. The distances from the bases of the isosceles trapezoid to the camera are set as $\underline{D}=30mm$ and $\overline{D}=80mm$, respectively.
				

The parameters of the improved wolf pack algorithm are set as shown in Table. \ref{parameter of IWPA}. A fair comparison of standard wolf pack algorithm and the improved wolf pack algorithm is also given with the same optimization parameters.

	\begin{table}[!t]
		\vspace{4pt}
		\centering
		\caption{Parameter of the IWPA in Simulation}
		\label{parameter of IWPA}
		\setlength{\tabcolsep}{2.2mm}{
			\begin{tabular}{c|c|c|c|c|c}
				\hline\hline
				\specialrule{0em}{0pt}{3pt}
		Parameter & Settings & Parameter  & Settings & Parameter  & Settings  \\
				\specialrule{0em}{0pt}{3pt}
				\hline
				\specialrule{0em}{0pt}{3pt}
				$Q$ & 25 & $L$ & 18 & $\Upsilon_{d}$ & 0.4  \\
				\specialrule{0em}{0pt}{3pt}
				\hline
				\specialrule{0em}{0pt}{3pt}
				$\Upsilon_{e}$ & 0.3 & $time_{a}$ & 6  & $\theta_{w}$ & $2^{\circ}$ \\
				\specialrule{0em}{0pt}{3pt}
				\hline
				\specialrule{0em}{0pt}{3pt}
				$step_{ap}$ & $3mm$ & $step_{ao}$ & $3^{\circ}$ & $G_{a}$ & 5\\
				\specialrule{0em}{0pt}{3pt}
				\hline
				\specialrule{0em}{0pt}{3pt}
				$\eta_{w}$ & $1mm$ & $time_{b}$ & 8 & $step_{bp}$ & $2mm$ \\
				\specialrule{0em}{0pt}{3pt}
				\hline
				\specialrule{0em}{0pt}{3pt}
				$step_{bo}$ & $2^{\circ}$ & $D_{be}$ & $3mm$ & $\lambda_{c}$ & $[-1,1]$ \\
				\specialrule{0em}{0pt}{3pt}
				\hline
				\specialrule{0em}{0pt}{3pt}
				$step_{cp}$ & $0.5mm$ & $step_{co}$ & $1^{\circ}$ & $time_{c}$ & 5 \\
				\specialrule{0em}{0pt}{3pt}
				\hline\hline
		\end{tabular}}
	\end{table}
	
\begin{figure*}[!t]
	\centering
	\begin{minipage}[t]{0.497\textwidth}
		\centering
		\includegraphics[width=8cm]{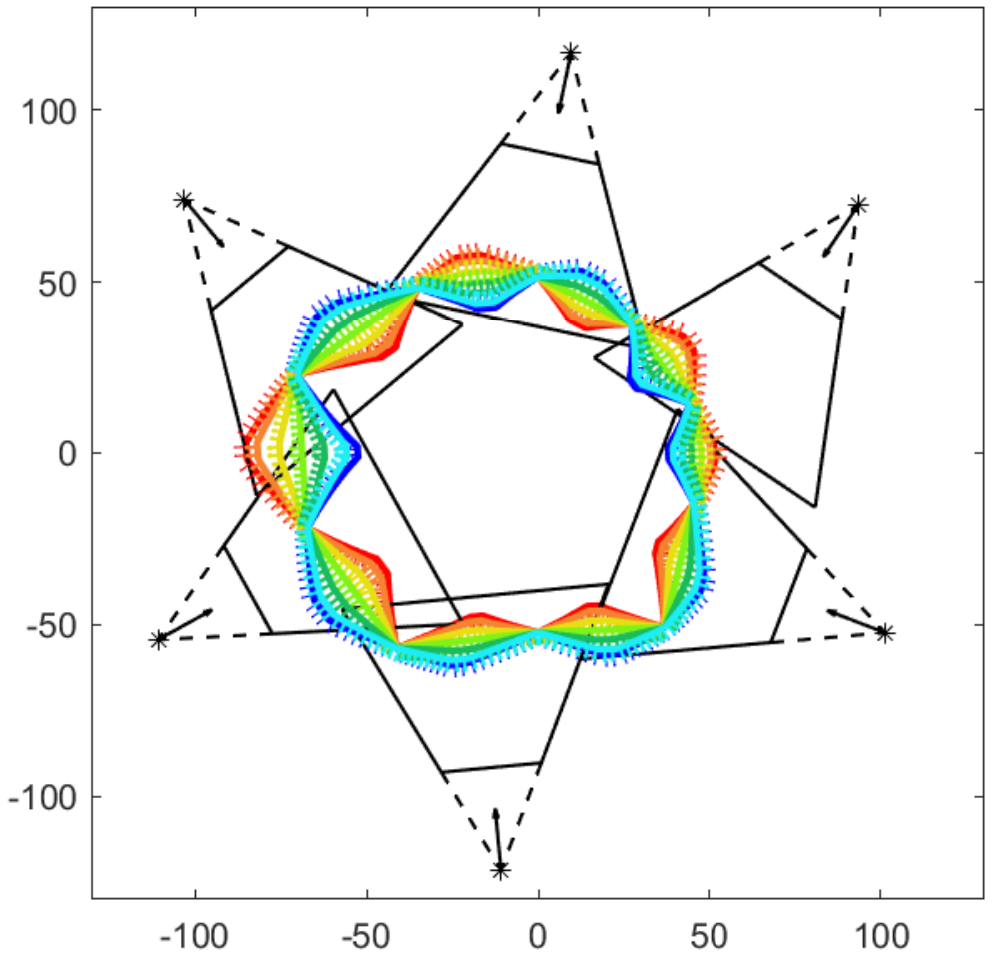}
		\caption{The final deployment of camera network under IWPA}
		\label{The final deployment}
	\end{minipage}
	\begin{minipage}[t]{0.497\textwidth}
		\centering
		\includegraphics[width=7.8cm]{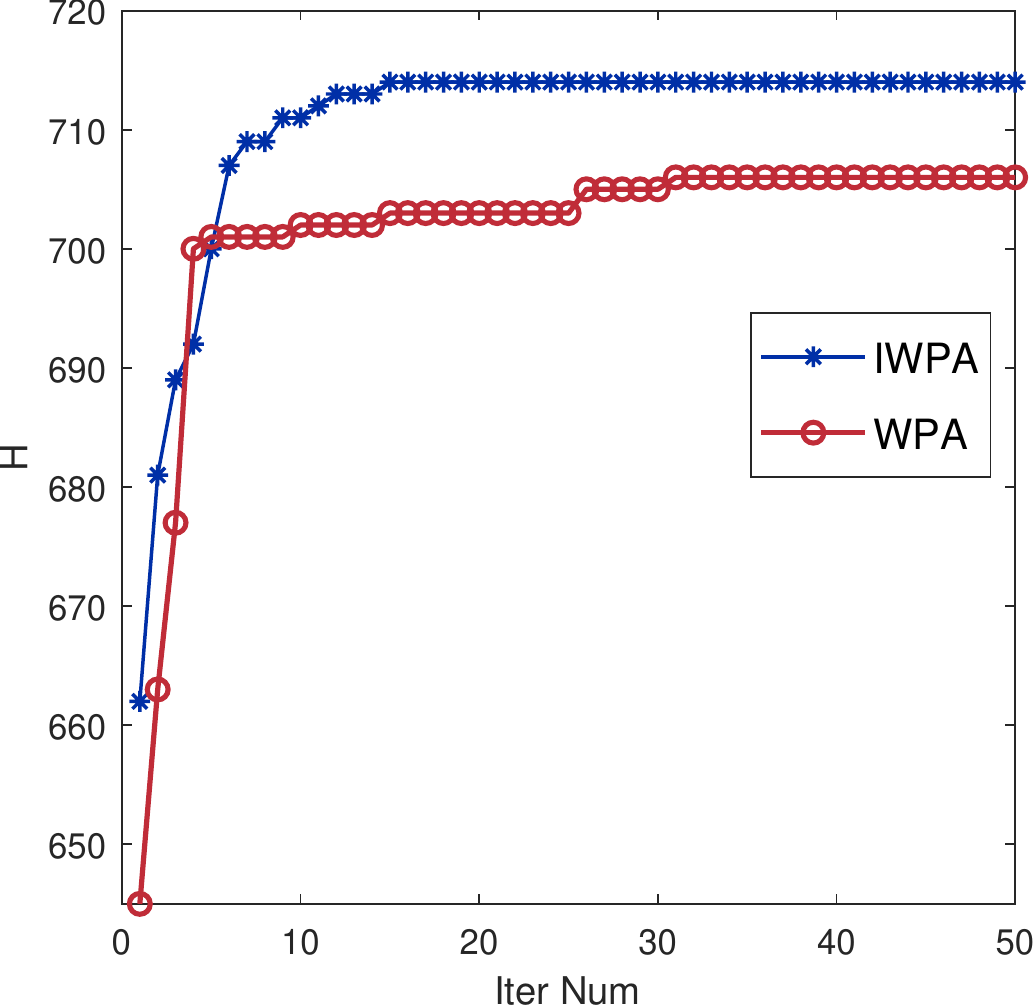}
		\caption{The cost function evolution}
		\label{The evolution of cost function}
	\end{minipage}
\end{figure*}

\begin{table*}[!t]
	\vspace{4pt}
	\centering
	\caption{Coverage rate of deformable contour at each instant}
	\label{coverage rate}
	\setlength{\tabcolsep}{3mm}{
		\begin{tabular}{lccccccccccc}
			\hline\hline
			\specialrule{0em}{0pt}{3pt}
			Contour & $\Omega_{t_1}$ & $\Omega_{t_2}$ & $\Omega_{t_3}$ & $\Omega_{t_4}$ & $\Omega_{t_5}$ & $\Omega_{t_6}$ & $\Omega_{t_7}$ & $\Omega_{t_8}$ & $\Omega_{t_9}$ & $\Omega_{t_{10}}$ & $\Omega_{t_{11}}$\\
			\specialrule{0em}{0pt}{3pt}
			\hline
			\specialrule{0em}{0pt}{3pt}
			Coverage Rate & 98.89\% & 98.83\% & 98.89\% & 99.44\% & 100\% & 100\% & 100\% & 100\% & 99.44\% & 100\% & 98.89\%\\
			\specialrule{0em}{0pt}{3pt}
			\hline\hline
	\end{tabular}}
\end{table*}

The simulation results are shown in Fig. \ref{The final deployment} and Fig. \ref{The evolution of cost function}. The final deployment of camera network under IWPA is shown in Fig.\ref{The final deployment}. It can be seen that the cameras are configured at better positions and orientations to achieve good coverage performance. The evolution of the cost function defined in (\ref{eq:cost function}) under IWPA and WPA are shown in Fig. \ref{The evolution of cost function}. We can see that the improved wolf pack algorithm has faster convergence speed and better performance. Define the coverage rate as the covered points by cameras and the total sampling points. The coverage rate of continuous deformable contour with the final deployment of cameras is shown in Table. \ref{coverage rate}, for some time instants on time interval $[t_0,t_s]$ (sampling per second). It can be seen that the coverage task of continuous deformable contour can be well accomplished through our proposed method.

\section{Conclusion}

In this paper, the deployment optimization of networked cameras for visual coverage on 2-D object with deformable contour is studied. The deformation process is formulated by continuous trajectories for the sampled points on the object contour. A rectangle is constructed for each trajectory representing the deformation process of the sampled point, by which the entire trajectory is contained. The vertexes of all the rectangles are selected as the feature points. The visual coverage task for deformable object is transferred to cover the selected feature points. An improved wolf pack algorithm is then proposed to solve the optimization problem.  At last, a simulation example is given to illustrate the effectiveness of the proposed method.




\bibliographystyle{IEEEtran}
\bibliography{IEEEabrv,IEEEexample}

\vfill

\end{document}